\documentclass{article}

\usepackage[preprint]{neurips_2025}

\usepackage[utf8]{inputenc} 
\usepackage[T1]{fontenc}    
\usepackage{hyperref}       
\usepackage{url}            
\usepackage{booktabs}       
\usepackage{amsfonts}       
\usepackage{nicefrac}       
\usepackage{microtype}      
\usepackage[table]{xcolor}         
\usepackage{graphicx}
\usepackage{multirow}
\usepackage{amsmath}
\usepackage{amssymb}
\usepackage{mathtools}
\usepackage{amsthm}
\usepackage{microtype}
\usepackage{graphicx}
\usepackage{subfigure}
\usepackage{multirow}
\usepackage{amssymb}
\usepackage{pifont}
\usepackage{array}
\usepackage{mathabx}
\usepackage{booktabs}
\usepackage{caption}
\usepackage{subcaption}
\usepackage{makecell}
\usepackage[most]{tcolorbox}
\usepackage{enumitem}

\usepackage{soul}
\usepackage{xcolor}
\definecolor{softred}{HTML}{FFE5E5}   
\definecolor{softgreen}{HTML}{E9F1E2} 

\definecolor{rowcolorblue}{RGB}{222, 232, 251}

\title{\Large AOR: Anatomical Ontology-Guided Reasoning for Medical \\ Large Multimodal Model in Chest X-Ray Interpretation}

\author{
Qingqiu Li$^1$\hspace{3mm} Zihang Cui$^2$\hspace{3mm} Seongsu Bae$^3$\hspace{3mm} Jilan Xu$^1$ \hspace{3mm}Runtian Yuan$^1$\hspace{3mm} Yuejie Zhang$^1$ \\ \textbf{Rui Feng}$^1$ \hspace{3mm} \textbf{Quanli Shen}$^4$ \hspace{3mm} \textbf{Xiaobo Zhang}$^4$ \hspace{3mm} \textbf{Junjun He}$^5$ \hspace{3mm} \textbf{Shujun Wang}$^6$\\
$^1$Fudan University\hspace{5mm}$^2$Xidian University\\$^3$KAIST\hspace{5mm}$^4$Children's Hospital of Fudan University\\$^5$Shanghai AI Laboratory\hspace{5mm}$^6$Hong Kong Polytechnic University \vspace{1mm}\\
\parbox{0.03\textwidth}{\includegraphics[width=\linewidth]{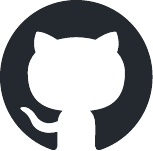}}\hspace{0.5mm}\textcolor{cyan}{\nolinkurl{https://aor-mllm.github.io/aor.html}}\vspace{-3mm}
}

\begin{document}

\maketitle

\begin{abstract}
\vspace{-2mm}
Chest X-rays (CXRs) are the most frequently performed imaging examinations in clinical settings. Recent advancements in Large Multimodal Models (LMMs) have enabled automated CXR interpretation, enhancing diagnostic accuracy and efficiency. However, despite their strong visual understanding, current Medical LMMs (MLMMs) still face two major challenges: (1) Insufficient region-level understanding and interaction, and (2) Limited accuracy and interpretability due to single-step reasoning.  
In this paper, we empower MLMMs with anatomy-centric reasoning capabilities to enhance their interactivity and explainability. 
Specifically, we first propose an Anatomical Ontology-Guided Reasoning (AOR) framework, which centers on cross-modal region-level information to facilitate multi-step reasoning. Next, under the guidance of expert physicians, we develop AOR-Instruction, a large instruction dataset for MLMMs training. Our experiments demonstrate \hspace{-0.1mm}AOR's \hspace{-0.1mm}superior \hspace{-0.1mm}performance \hspace{-0.1mm}in \hspace{-0.1mm}both \hspace{-0.1mm}VQA \hspace{-0.1mm}and \hspace{-0.1mm}report \hspace{-0.1mm}generation \hspace{-0.1mm}tasks.
\end{abstract}

\vspace{-4mm}
\section{Introduction}
\label{sec:intro}
\vspace{-2mm}

Large Language Models (LLMs) have demonstrated impressive capabilities in knowledge and reasoning~\cite{gpt4, llama1, llama1.5}. This advancement has inspired the community to extend LLMs' foundational abilities to the visual domain, leading to the development of Large Multimodal Models (LMMs)~\cite{llava, llava1.5}. In the medical field, LMMs are also gaining increasing attention, with prominent models such as LLaVA-Med~\cite{llavamed} and CheXagent~\cite{chexagent}. By integrating medical visual and textual modalities, LMMs can facilitate various medical needs, including patient consultation, medical report generation, and disease diagnosis~\cite{foundation}.

\begin{figure}[t]
\centering
\includegraphics[width=\linewidth]{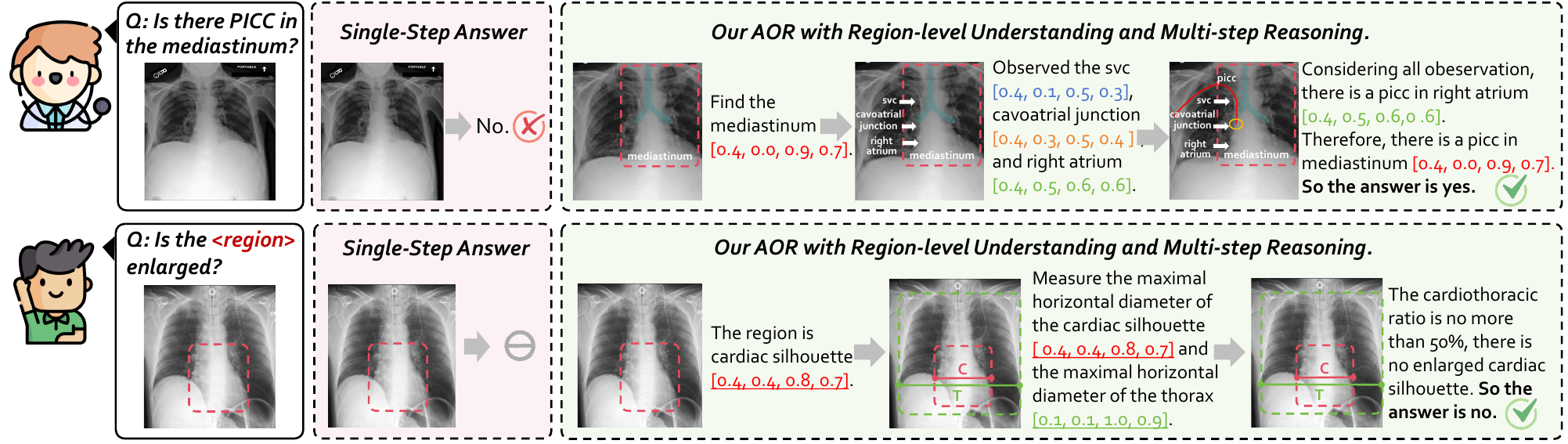}
\caption{
Previous image-level MLMMs (shown in {\sethlcolor{softred}\hl{light red}}) make incorrect predictions or fail to predict due to (1) insufficient region-level perception and (2) reliance on single-step reasoning. 
In contrast, our AOR model (shown in {\sethlcolor{softgreen}\hl{light green}}) delivers explainable and accurate answers by (1) emphasizing region-level understanding and (2) employing multi-step reasoning.
}
\label{fig:introduction}
\vspace{-2mm}
\end{figure}

While existing Medical LMMs (MLMMs) have demonstrated remarkable capabilities in visual understanding, they still encounter two significant challenges that limit their effectiveness in medical imaging applications:

\textbf{(1) Insufficient Region-Level Understanding and Interaction.} 
Radiologists analyze CXR by first assessing the overall image and then focusing on specific anatomical regions to identify abnormalities. To replicate this process, models must understand visual details, spatial relationships, and hierarchical anatomical structures. However, current image-level MLMMs~\cite{llavamed, xraygpt} struggle to detect subtle, clinically significant lesions. Additionally, as shown in Fig.~\ref{fig:introduction}, region-level interaction is crucial: radiologists need to revisit areas of interest, while non-expert users rely on models to interpret visual cues without precise medical terminology. Existing MLMMs lack these capabilities, limiting their real-world applicability.
Thus, it is imperative to highlight region-level perception to
overcome these obstacles.

\textbf{(2) Limited Accuracy and Interpretability due to Single-Step Reasoning}. 
The complexity of medical imaging analysis stems from overlapping symptoms across diseases and diverse manifestations of the same condition, thus requiring multi-step reasoning for accurate diagnoses. 
Effective models must integrate CXR images with clinical questions, analyzing lesion location, characteristics, and contextual associations. 
However, current MLMMs rely solely on single-step, black-box reasoning, leading to inaccuracies, misinterpretations~\cite{chan2}, and hallucinations~\cite{hallu}, as they fail to capture nuanced symptom-lesion-disease relationships. However, creating high-quality instruction data for multi-step reasoning is challenging due to the specialized nature and low error tolerance of medical imaging. 
These limitations hinder the development of accurate, interpretable models essential for reliable medical imaging analysis. Therefore, creating a clinically credible Chain of Thoughts (CoT) dataset is vital to augmenting the reasoning performance of MLMMs.

In this paper, we propose the Anatomical Ontology-Guided Reasoning (AOR) framework with a three-stage training strategy. AOR centers on the anatomical regions relevant to the given question, incorporating their positional and representational information to conduct multimodal multi-step reasoning. Then, to address the shortage of multimodal reasoning datasets for MLMMs, we develop AOR-Instruction dataset under the guidance of expert physicians, consisting of two subsets: AOR-VQA and AOR-RG. Specifically, for AOR-VQA, we construct 2,812 CoT templates under three anatomical ontologies to provide precise CoT answers for 290k VQA samples. For AOR-RG, 133k CXR-report pairs are used for full image report generation, while raw reports are decomposed into fine-grained descriptions, forming 399k strictly aligned region-sentence pairs for interpretable region report generation.

By empowering Medical LMMs with anatomy-centric reasoning capabilities, we offer a new paradigm for interactive and explainable LMMs in medical imaging analysis. The contributions of this work are summarized as:

\begin{itemize}[leftmargin=*]
    \item We propose an Anatomical Ontology-Guided Reasoning (AOR) framework. It supports both textual and optional visual prompts as input, centered on region-level information to enable multimodal multi-step reasoning.
    \item We develop a large instruction dataset named AOR-Instruction by leveraging anatomical regions and their ontologies. It consists of two parts: AOR-VQA for Visual Question Answering (VQA) and AOR-RG for full image and region report generation, containing 290k and 532k data pairs.
    \item Extensive experiments demonstrate the superiority of AOR, which outperforms the second-best MLMM by an average of 6.81\% on the VQA and 5.27\% on report generation, underscoring the crucial role of region perception and reasoning capabilities in supporting clinical decision-making.
\end{itemize}

\section{Related Works}
\label{sec:related}

\textbf{Medical Large Multimodal Models} With the success of LLMs~\cite{gpt4, llama1, llama1.5}, researchers are enhancing these models by incorporating visual understanding capabilities, leading to the emergence of LMMs~\cite{llava, llava1.5}. In the medical domain, numerous LMM-based studies have also arisen. Prominent models such as LLaVA-Med~\cite{llavamed} and Med-Flamingo~\cite{Med-flamingo} first perform image-text feature alignment using paired medical data, followed by meticulously designed instruction tuning. Although these models exhibit strong visual understanding, they are primarily limited to image-level tasks like report generation and medical visual question answering. They do not explicitly learn region-level features during the training process, which constrains their region-level perception.

\noindent\textbf{Region-Level Medical LMMs} To achieve more fine-grained image understanding, recent research has further integrated region-level data into the training of LMMs. Shikra~\cite{shikra} directly quantizes bounding boxes into coordinates (numerical representations of positions). Subsequently, GPT4RoI~\cite{gpt4roi} and RegionGPT~\cite{regiongpt} extract region features from the original images and include them as part of the input token sequences, allowing the models to fully comprehend region representations and enabling them to process visual prompts. However, in the medical field, research on region-level LMMs is still limited. BiRD~\cite{bird} aims to equip MLMMs with grounding and referring capabilities through multi-task learning while maintaining their core conversational ability. MAIRA-2~\cite{maira} focuses on improving the performance of LMMs in grounded report generation tasks. Both methods locate specific regions using textual coordinates and rely on single-step diagnoses, lacking the comprehensive perception and reasoning to fully utilize these regions.

\textbf{CoT in Medical LMMs} Chain of Thoughts (CoT) is a series of prompting strategies aimed at helping large language models (LLMs) address complex problems by guiding them through intermediate reasoning steps. Previous studies~\cite{llmcot1, llmcot2} have shown that LLMs benefit from carefully crafted CoT prompts. Recently, CoT has been increasingly integrated into Large Multimodal Models (LMMs). SoM~\cite{som} integrates supplementary information into images, e.g., segmentation maps. VoCoT~\cite{vocot} and Visual-CoT~\cite{visualcot} introduce CoT during training by constructing an instruction dataset to facilitate LMMs in adapting to object-centric reasoning. However, the introduction of CoT into medical LMMs remains largely underexplored. MedCoT~\cite{medcot} leverages Gemini-Pro~\cite{gemini} to assist in generating CoT. However, constructing the CoT process in the medical domain requires highly specialized domain knowledge, rather than relying entirely on LMMs.

\section{Method}

\subsection{Model Overview}
As illustrated in Fig.~\ref{fig:framework}, AOR mainly consists of three components: (i) an image encoder $\mathcal{I}$, responsible for extracting image features; (ii) a region encoder $\mathcal{R}$, deployed to extract multi-scale region features from image features; and (iii) a large language model $\mathcal{LLM}$ is designed to jointly model image, region, and text for reasoning after projecting image and region features into the linguistic space.

\subsection{Model Development}
\begin{figure}[t]
\centering
\includegraphics[width=\linewidth]{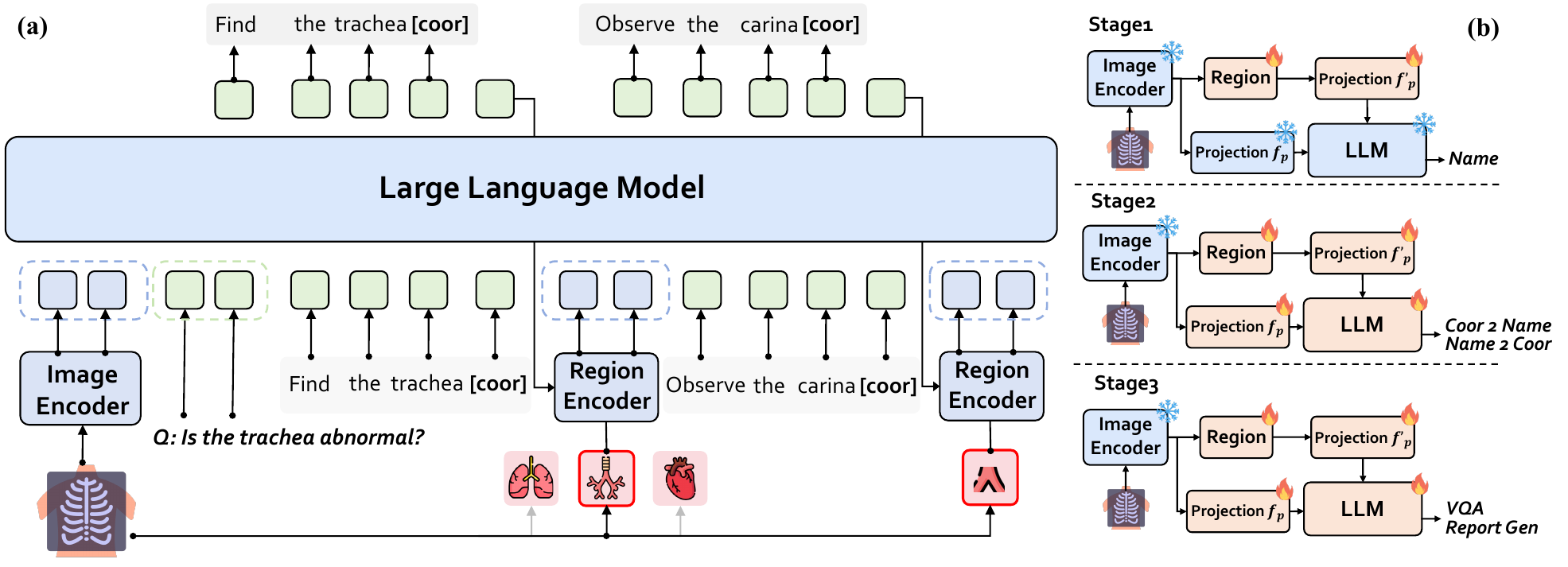}
\vspace{-2mm}
\caption{(a) Overview of AOR framework, which flexibly accommodates both textual and optional visual prompts as input, centered on region-level information to enable multimodal multi-step reasoning and (b) Three-stage training procedure for AOR.}
\label{fig:framework}
\vspace{-2mm}
\end{figure}

Fig.~\ref{fig:framework} (b) shows our three-stage training procedure for AOR. We progressively enabling AOR to perform anatomy-centric recognition, detection, reasoning, and report generation.

\textbf{Stage 1: Anatomical Region Recognition} The first stage aims to align region features with linguistic embeddings, enabling the model to recognize each anatomical region in CXR. During this stage, only the region encoder $\mathcal{R}$ and the region projection $f_{p}'$ are kept trainable. 
For each image $I$, we use the anatomical bounding boxes $B=\{(c^j, n^j)^{N_b}_{j=1}\}$ provided by Chest ImaGenome Dataset~\cite{chestImagenome}. Here, $c^j \in \mathbb{R}^4$, $n^j$ and $N_b$ represent the coordinate, region name, and the number of anatomical regions in $I$. 
The image $I$ is first encoded into the feature maps $\textbf{z}=\{z_i\}_{i=1}^{N_l}$, where $N_l$ is the number of feature maps.
Inspired by GPT4RoI~\cite{gpt4roi}, we design the region encoder $\mathcal{R}$ which constructs a hierarchical feature pyramid from four selected layers of the image encoder. According to $c^j$, RoIAlign is then applied to generate a 14$\times$14 feature map from $\textbf{z}$, followed by a pooling layer to embed multi-scale region features $r^j$. 
Image projection $f_{p}$ and region projection $f_{p}'$ are used to connect $z_{N_l}$ and $r^j$ into the linguistic space. 
Finally, the $\mathcal{LLM}$ integrates the projected visual features and the text instruction embedding $t$ to recognize the current anatomical region and output its name $n^j$:
\begin{equation}
    n^j = \mathcal{LLM}(f_p(z_{N_l}), t, f_p'(r^j))
    \vspace{-1mm}
\end{equation}

\textbf{Stage 2: Anatomical Region Grounding} In the second stage, the model is trained to localize anatomical regions, laying the foundation for subsequent reasoning tasks. Since the generation of coordinates requires an overview of the entire image and the generative capability of the LLM, we keep both the LLM and image projection modules trainable. Two types of tasks are considered: 

(1) To prevent catastrophic forgetting and align the format closer to the reasoning tasks, the model revisits the anatomical region recognition of Stage 1 with some adjustments, i.e., concatenating the coordinates to the region feature. Here, we use bbox [$x_{min}, y_{min}, x_{max}, y_{max}$] as object coordinates, where $x$ and $y$ are normalized between 0 and 1 relative to the image size. The LLM reads the projected visual features, textual coordinates embedding $c^j$, and text instruction embedding $t$ to predict the region name $n^j$:
\begin{equation}
    n^j = \mathcal{LLM}(f_p(z_{N_l}), t, [c^j,f'_p(r^j)])
    \vspace{-2mm}
\end{equation}

(2) Given region's name $n^j$, the model grounds the corresponding coordinates $c^j$:
\begin{equation}
    c^j = \mathcal{LLM}(f_p(z_{N_l}), t, n^j)
    \vspace{-1mm}
\end{equation}

\textbf{Stage 3: Instruction Tuning}
Based on the pre-trained model, this stage fine-tunes the model using AOR-Instruction (detailed in Section~\ref{instruction_data}) on three tasks:

(1) Medical Visual Question Answering: AOR is capable of handling questions that require both global and local clues, and is flexible enough to accept both textual and optional visual prompts as input. Based on the given prompt, the model centers on anatomical regions to generate logically reasoned answers. During reasoning, each region is represented in a triplet format: ⟨region name⟩ ⟨coordinates⟩ ⟨ROI visual representation⟩, e.g., ``svc [0.27, 0.08, 0.92, 0.81] $r_{svc}$''. Once the end of the coordinates token ``]'' is generated, the region encoder $\mathcal{R}$ is activated to obtain the ⟨ROI visual representation⟩ based on the coordinates between ``['' and ``]'', which is formulated as follows:
\begin{equation}
    ans^j = \mathcal{LLM}(f_p(z_{N_l}), t, [n^j, c^j, f_p'(r^j)])
    \vspace{-1.5mm}
\end{equation}

(2) Full Image Report Generation: Given a CXR, AOR generates a comprehensive report describing the entire image:
\begin{equation}
    report = \mathcal{LLM}(f_p(z_{N_l}), t)
    \vspace{-1mm}
\end{equation}

(3) Region Report Generation: For a chest X-ray, users provide textual and optional visual prompts specifying the anatomical regions of interest. AOR generates a report sentence $s^j$ specifically related to the designated region $r^j$.
\begin{equation}
    \vspace{-1mm}
    s^j = \mathcal{LLM}(f_p(z_{N_l}), t, f_p'(r^j))
\end{equation}

\section{Instruction Data}
\label{instruction_data}

\begin{figure}[t]
\centering
\includegraphics[width=\linewidth]{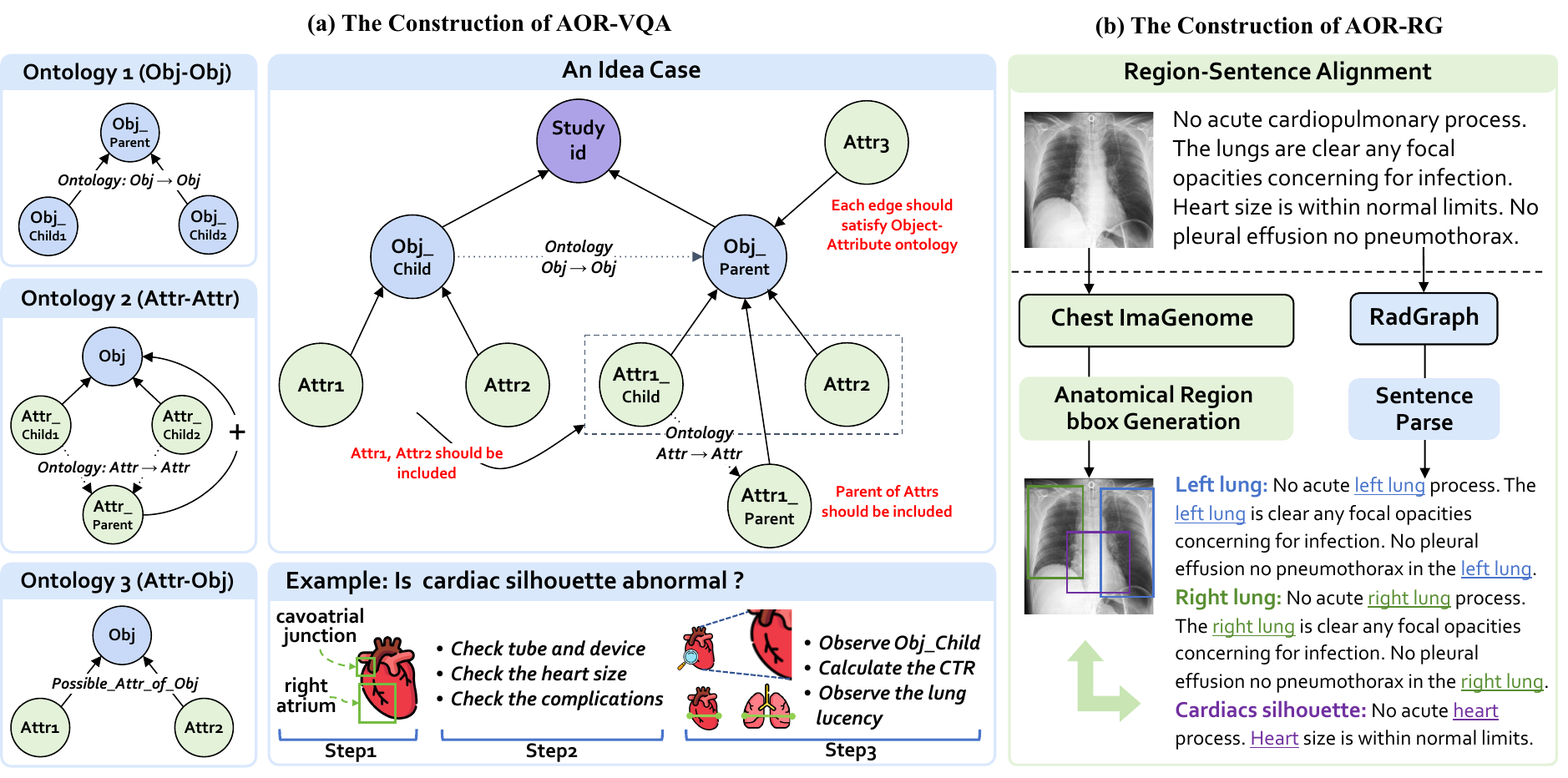}
\vspace{-5mm}
\caption{Overview of AOR-Instruction, which consists of two sub-datasets: (a) The construction of AOR-VQA: Anatomical ontologies design $\rightarrow$ CoT construction $\rightarrow$ Sample expansion and (b) The construction of AOR-RG: Strict alignment between anatomical region and report sentence.}
\label{fig:data}
\vspace{-3mm}
\end{figure}

Currently, there is a shortage of multimodal reasoning datasets for training Medical LMMs, leading to models that lack fine-grained understanding and reasoning capabilities. To bridge this gap, we construct the AOR-Instruction dataset by leveraging anatomical regions and their ontologies. This dataset, enriched with explainable region-level visual information, helps LMMs achieve a clearer understanding of image content. The AOR-Instruction dataset consists of two parts: AOR-VQA for Visual Question Answering (VQA) and AOR-RG for full image and region report generation, containing 290k and 532k data pairs, respectively.

\subsection{AOR-VQA}
AOR-VQA is primarily designed to enhance the model’s
capabilities in medical VQA. We use MIMIC-CXR-VQA~\cite{ehrxqa}, a comprehensive dataset with (Image, Question, Answer) samples, as our primary data source. It includes both global-level questions that evaluate the overall findings of a CXR (e.g., ``Are there any abnormalities?''), and local-level questions that focus on specific anatomical regions (e.g., ``Is there a nodule in the left lung?'').

The entire construction process is conducted under the guidance of expert physicians—two board-certified radiologists and one clinician with 24, 18, and 27 years of experience, respectively. We meticulously design and refine three types of anatomical ontologies. Based on these ontologies, we construct a Chain of Thought (CoT) for each sample. Finally, the expanded information is attached to each sample, resulting in a structure that includes (Image, Question, Region Box, CoT Answer). The details are as follows:

\subsubsection{Anatomical Ontologies Design}

\indent We define anatomical regions in the CXR as objects, each associated with several attributes selected from a pool of 68 attributes across five categories. Fig.~\ref{fig:data} (a) illustrates our anatomical ontologies. 

\textbf{Ontology 1: Hierarchical Relationships Between Objects}
In visual perception, humans organize content into hierarchical structures to understand the part-whole relationships within images, thereby obtaining the answers they seek. Fortunately, such part-whole hierarchical relationships clearly exist between anatomical regions. As illustrated in Fig.~\ref{fig:data} (a), Obj\_Parent (e.g., mediastinum) includes two related objects: Obj\_child1 (e.g., upper mediastinum) and Obj\_child2 (e.g., cardiac silhouette). By leveraging this hierarchical relationship, the model can engage in reasoning by shifting focus from the whole to the parts and then comprehensively considering the whole based on the parts. Therefore, we explore and organize the hierarchical relationships of 38 anatomical regions. 

\textbf{Ontology 2: Causal Relationships Between Attributes}
During the progression of specific attributes, multiple conditions can interrelate. Additionally, attributes categorized as anatomical findings are typically the imaging manifestations of attributes classified as diseases. Thus, we can construct causal relationships between different attributes. As shown in Fig.~\ref{fig:data} (a), if Attr\_Child (e.g., lobar/segmental collapse) exists, then Attr\_Parent (e.g., atelectasis) within an Obj must also be present. Leveraging these causal relationships can help the model understand the associations between attributes and utilize other attributes to complete the reasoning process when encountering attributes with unclear or difficult-to-determine visual manifestations. 
Consequently, the causal relationships of 68 attributes are constructed.

\textbf{Ontology 3: Restrictive Relationships Between Objects and Attributes} Finally, we consider the restrictive relationships between objects and attributes. As shown in Fig.~\ref{fig:data} (a), only certain attributes appear within specific objects. For example, fractures can never occur at the cardiac silhouette, while pleural effusion most commonly affects the costophrenic angle. By utilizing such restrictive relationships, the model can eliminate certain scenarios, enabling more rational and effortless reasoning. Therefore, the restrictive relationships between 38 objects and 68 attributes are organized.

\subsubsection{CoT construction}

\indent Integrating the aforementioned three ontologies, as illustrated in Fig.~\ref{fig:data} (a), we organize an ideal case for each image (represented as a Study id node) in the source dataset, establishing connections between all objects and attributes for each sample. For global-based questions, we do not expand the answers, enabling the model to make fine-grained observations while also developing global summarization skills. For local-based questions, we construct a rigorous and comprehensive CoT for each question based on the ideal case, following the steps below.

\textbf{Step 1: Identify Sub-objects (Using Ontology 1)} For the queried object, we first identify its sub-objects (if it is already the smallest anatomical structure unit, this step is skipped). Considering the question: ``Is the cardiac silhouette abnormal?'', we identify the sub-objects of the cardiac silhouette as the right atrium and the cavoatrial junction.

\textbf{Step 2: Consider all possible attributes (Using Ontology 2)} If the question targets a specific attribute, analysis and reasoning are conducted solely for that attribute. However, if the question concerns all abnormalities of the queried object, all possible scenarios must be considered. Continuing with the example from Step 1, for abnormalities in the cardiac silhouette, this primarily includes the presence of tubes or devices, changes in the size of the cardiac silhouette, and the development of other complications.

\textbf{Step 3: Associate the relevant objects and attributes (Using Ontology 3)} Once we have identified the attributes to be discussed, we focus our observation and reasoning on its primary associated object or sub-objects. Continuing with the example above: for the presence of tubes or devices, particularly observe the right atrium and the cavoatrial junction; for measurements, i.e., cardiac silhouette size, place the cardiac silhouette within the global context and compare it to the size of the entire thorax; for the development of complications, after detecting an enlarged cardiac silhouette, consider other features, i.e., pulmonary translucency, to further assess the presence of lung opacity.

Following the three steps outlined above, we can construct a CoT answer for any combination of objects and attributes (or categories and abnormalities). A total of \textbf{(68$+$5$+$1)$\times$38 $=$ 2,812} types of CoT answers are constructed, all of which are reviewed and refined by three expert physicians.

\indent Furthermore, source data includes complex combinatorial questions, such as those involving conjunction and disjunction, which are commonly seen in real-world application scenarios. Therefore, we decouple such data by extracting the involved sub-questions to perform the aforementioned CoT and finally conduct an additional logical inference based on the answers to the sub-questions.

\subsubsection{Sample Expansion}
\vspace{-1mm}
All samples in the dataset are expanded from (Image, Question, Answer) to (Image, Question, Region Box, CoT Answer) according to above rules.

\subsection{AOR-RG}
\textbf{Full image report generation} We directly utilize the image-report pairs provided in MIMIC-CXR~\cite{mimic-cxr}, make use of frontal images, and include findings and impressions in the report.

\textbf{Region report generation} To perform fine-grained region report generation, we need to decompose raw report data into fine-grained descriptions for each organ mentioned in medical scans. As shown in Fig.~\ref{fig:data} (b), we utilize the bounding boxes provided by Chest ImaGenome Dataset and parsed the text using RadGraph~\cite{radgraph}, employing the rules proposed in ASG~\cite{asg} to achieve strict alignment between the two. Additionally, we further optimize the alignment method to handle cases where two different anatomical regions appear in the same short sentence, introducing new rules to split such sentences into two separate ones. This approach constructs region-sentence pairs for each image-report pair.

\section{Experiments}
\subsection{Experiment Settings}
\begin{table}[]
\centering
\caption{Comparison of methods on MIMIC‑CXR‑VQA, VQA‑RAD, and CheXpert. ``–'' means Med‑Flamingo lacks training code, so it cannot be fine‑tuned on MIMIC‑CXR‑VQA. ``*'' means CheXagent likewise lacks training code, but its instruction data include MIMIC‑CXR‑VQA, so we report zero‑shot results. Text in \textcolor{gray}{gray} shows scores where CheXagent data include VQA‑RAD, so they are excluded from comparison. \textbf{Bold} numbers mark the best result in each column.}
\vspace{3mm}
\begin{tabular}{lcccccccc}
\toprule
\multirow{2}{*}{\textbf{Method}} & \multirow{2}{*}{Res.} & \multicolumn{3}{c}{MIMIC-CXR-VQA} & \multicolumn{2}{c}{VQA-RAD} & \multicolumn{2}{c}{CheXpert} \\ \cmidrule(lr){3-5} \cmidrule(lr){6-7} \cmidrule(lr){8-9}
&    & verify  & choose   & query  & closed & open & closed  & open       \\
\cmidrule(r){1-2} \cmidrule(lr){3-5} \cmidrule(lr){6-7} \cmidrule(lr){8-9}
\multicolumn{9}{l}{{\textit{\textbf{General-domain LMM}}}}\\
LLaVA~\cite{llava}                           & $\text{224}^2$      &75.97  &56.07 &58.87 &43.84  &21.09  &27.72  &34.50        \\
LLaVA-1.5~\cite{llava1.5}                      & $\text{336}^2$    &75.25  &\underline{58.70} &56.10 &44.74  &13.54  &27.72  &33.88         \\
GPT4RoI~\cite{gpt4roi}                         & $\text{224}^2$    &\underline{77.16}  &56.37 &60.54 &43.84  &17.64  &52.19  &32.65         \\
VoCoT~\cite{vocot}                          & $\text{448}^2$       &76.17  &48.79 &\underline{60.70} &35.62  &21.38  &49.60  &40.89    \\
\cmidrule(){1-9}
\multicolumn{9}{l}{{\textit{\textbf{Medical-domain LMM}}}} \\
LLaVA-Med~\cite{llavamed}                       &$\text{224}^2$    &75.71  &58.31  &60.37  &\underline{61.64} &22.36 &54.95  &\underline{42.08}    \\
Med-Flamingo~\cite{Med-flamingo}                & $\text{224}^2$   & -    & -      & -     &28.77 &\underline{23.89} &40.99  &39.80    \\
XrayGPT~\cite{xraygpt}                        & $\text{224}^2$     &60.00  &40.97  &24.07  &43.84 &22.51 &60.59  &30.45   \\
CheXagent~\cite{chexagent}                       & $\text{448}^2$  &75.02*  &33.49*  &48.49*  &\textcolor{gray}{68.49} &\textcolor{gray}{24.94} &\underline{62.28}  &32.14         \\ \hline
\rowcolor{rowcolorblue}
AOR(Ours)-t                      & $\text{336}^2$                   & \textbf{80.48}  & \textbf{71.96}  & \textbf{65.05}  & \textbf{63.01}  
& \textbf{28.19}         & \textbf{71.58}      & \textbf{53.85}         \\
\rowcolor{rowcolorblue}
AOR(Ours)-r/t                    & $\text{336}^2$                   & 80.68       & 70.16      & 65.43     &57.53  & 24.99    &  74.06      & 45.35      \\ \bottomrule
\label{tab: tab1}
\end{tabular}
\vspace{-6mm}
\end{table}

\textbf{Implementation Details} We initialize the image encoder with CLIP-ViT-L/14~\cite{clip} and the language model with LLaVA-1.5~\cite{llava1.5}. The input resolution for images is set to 336$\times$336. We use AdamW as our optimizer, with a learning rate of $\text{2}\times\text{10}^{\text{-5}}$. Experiments are conducted using 4 NVIDIA A100 GPUs.

\textbf{Dataset} 1) For the VQA task, we evaluate on the test sets of MIMIC-CXR-VQA~\cite{ehrxqa}, VQA-RAD~\cite{vqarad}, and CheXpert~\cite{chexpert}. MIMIC-CXR-VQA contains 500 images and 13,793 QA pairs in the test dataset. CheXpert contains 191,229 frontal chest radiographs, and we hold out the expert-labeled validation set as the test data, which contains 202 images and 1212 QA pairs. VQA-RAD includes 315 images and 3,515 QA pairs distributed across the head, chest, and abdomen. We filter it to include only chest X-ray images and their corresponding question-answer pairs, resulting in 69 images and 102 QA pairs in the test dataset. 2) For the full image report generation task, we use MIMIC-CXR, a large publicly available dataset of chest radiographs with free-text radiology reports, to evaluate the model's performance. The test dataset contains 500 images and their corresponding reports. 3) For the region report generation task, we use the same 500 images and sample 3 anatomical regions per image for evaluation.

\begin{figure}[t]
  \centering
\begin{minipage}[t]{0.48\linewidth}
\centering
\captionof{table}{Comparison of methods on MIMIC-CXR dataset.}
\setlength{\tabcolsep}{4pt}
\renewcommand{\arraystretch}{1.2}
\resizebox{\linewidth}{!}{
\begin{tabular}{lccc}
\toprule
\multirow{2}{*}{\textbf{Method}} & \multicolumn{3}{c}{MIMIC-CXR}         \\ \cline{2-4} 
& \ \ \ R-L\ \ \  & BERTScore & F\textsubscript{1}CheXbert  \\ \hline
\multicolumn{4}{@{}l}{\textit{\textbf{Full image report generation}}}                      \\
LLaVA-Med   & 13.49 & 75.93    & 0.40  \\
Med-Flamingo & 5.21 & 71.26    & 13.61   \\
XrayGPT      & 24.02 & 83.18   & 26.71   \\
CheXagent   & \underline{24.09} & \underline{83.52}    & \underline{37.54}          \\ \hline
\rowcolor{rowcolorblue}
AOR(Ours)-t  & \textbf{25.37}    & \textbf{83.92}     & \textbf{46.53}\\  
\rowcolor{rowcolorblue}
AOR(Ours)-r/t  & 25.38 &83.95 & 48.28              \\ \hline
\multicolumn{4}{@{}l}{\textit{\textbf{Region report generation}}}                          \\
LLaVA-Med    & 9.47  & 71.51 & 16.70         \\
Med-Flamingo  & 12.98 & 75.24  & 14.66   \\
XrayGPT      & 15.80 & 80.19  & 19.96                  \\
CheXagent    & \underline{20.79}             & \underline{81.57}  & \underline{33.02}  \\ \hline
\rowcolor{rowcolorblue}
AOR(Ours)-t  & \textbf{35.11}  & \textbf{84.54}  & \textbf{36.65}       \\ 
\rowcolor{rowcolorblue}
AOR(Ours)-r/t & 35.62  & 84.76  & 36.89      \\ \bottomrule
\end{tabular}
}
\label{tab:rg comparison}
\end{minipage}
\hfill
\begin{minipage}[t]{0.48\linewidth}
\centering
\captionof{table}{Comparison between CoT in different formats.}
\setlength{\tabcolsep}{4pt}
\resizebox{\linewidth}{!}{
\begin{tabular}{ccccccc}
\toprule
\textbf{ID} & \textbf{coor} & \textbf{region} & \textbf{CoT} & verify & choose & query \\ \hline
1  & \ding{55}  & \ding{55}  & \ding{55}  & 76.83 & 61.07 & 58.77 \\
2  &  \ding{55} & \ding{55}  & \checkmark & \textbf{80.69} & 67.62 & 63.37 \\
3  & \checkmark & \ding{55}  & \checkmark & 79.14 & \underline{69.54} & \underline{63.89} \\
4  & \checkmark & \checkmark & \checkmark & \underline{80.68} & \textbf{70.16} & \textbf{65.43} \\ \bottomrule
\end{tabular}
}
\label{tab:cot comparison}

\vspace{18pt}

\centering
\resizebox{\linewidth}{!}{
\includegraphics[width=1\linewidth]{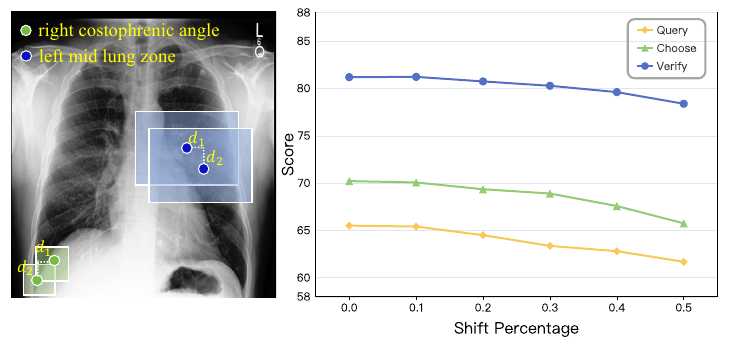}
}
\captionof{figure}{Impact of anatomical region shifts on model predictions.}
\label{fig:exp}
\end{minipage}

\end{figure}

\textbf{Evaluation metrics.} For the VQA task, regarding MIMIC-CXR-VQA, its questions can be categorized into three primary semantic types: For the ``verify'' questions, which include yes/no questions, we report the accuracy; for the ``choose'' questions, which involve selection from provided options, we also report the accuracy; for the ``query'' questions, where the answers are in the form of a list, we report the F\textsubscript{1} score (micro).
For CheXpert and VQA-RAD, following \cite{llavamed}, for closed-end questions with a single correct answer, we report accuracy; for open-end questions, we use recall to evaluate the model's responses. For the report generation task, we selected ROUGE-L (R-L)~\cite{lin2004rouge}, BERTScore~\cite{zhang2019bertscore}, and F\textsubscript{1}CheXbert~\cite{miura2021improving} 
as evaluation metrics to compare model performance at the word level, semantic level, and clinical efficacy level.

\subsection{Quantitative Comparison}
\textbf{Performance on MIMIC-CXR-VQA} 
We fine-tune all models on MIMIC-CXR-VQA. For a fair comparison, we introduce AOR-t, which uses the same training data as the baseline methods, where all questions are provided in textual form. AOR-r/t represents a setting where questions are presented in both textual and visual formats during training, enabling multimodal interaction. As revealed in Table~\ref{tab: tab1}, AOR-t outperforms the second-best method by an average of 6.98\%, especially on the more complex ``choose'' and ``query'' question types, highlighting the advantage of reasoning centered on anatomical regions. 

\textbf{Zero-shot transfer to VQA-RAD and CheXpert} We evaluate AOR’s generalization ability on unseen data distributions, i.e., VQA-RAD and CheXpert. As shown in Table~\ref{tab: tab1}, AOR-t obtains superior performance on both datasets. Moreover, AOR is capable of reasoning and providing logical answers rather than simply responding with ``yes'' or ``no''.

\textbf{Performance on MIMIC-CXR} Table~\ref{tab:rg comparison} shows that AOR outperforms all previous methods in full-image and region-based report generation. Notably, it addresses the gap in generating report sentences for specific regions, which is a limitation of previous MLMMs.

\subsection{Ablation Studies and Discussions}
\textbf{Comparison between CoT in Different Formats} As shown in Table~\ref{tab:cot comparison}, we demonstrate the effectiveness of the AOR format by comparing it with different CoT formats for the VQA task. ID-1: The original answer is used without the CoT process, whose accuracy is limited, especially for more complex questions. ID-2: The CoT answer represents objects using only text descriptions. The inclusion of CoT significantly improves the model's performance, highlighting the importance of multi-step reasoning. ID-3: The CoT answer augments ID-2 with textual coordinates, enabling the model with spatial awareness to first locate key anatomical regions and then further observe them. ID-4: Our AOR format, builds upon ID-3 by cropping and embedding multi-scale region features. This helps the model utilize the visual cues of objects, providing the best performance.

\textbf{Referring and Grounding Capabilities} We analyze the model’s referring and grounding capabilities to valid the rationality of the CoT process. In Table \ref{tab:diff stage}, under the setting where all three training stages are involved, the model achieves a 98.58\% referring accuracy and 90.40\% R@0.7, ensuring both the accuracy and explainability of the answers. When Stage 2 is removed, the recall drops accordingly, indicating that the grounding task in Stage 2 lays a solid foundation for reasoning. Similarly, without the process of Stage 1, the performance of referring is affected, demonstrating the effectiveness of our three-stage training. 

\textbf{The Impact of Anatomical Region Shifts on Model Predictions} In practical applications, considering that radiologists may not provide perfectly accurate region prompts and patients might offer bboxes that deviate from standard anatomical regions, we explore the impact of bbox coordinate shifts on prediction accuracy. As shown in Fig.~\ref{fig:exp}, we apply shifts to the bbox in the horizontal and vertical directions during model inference ($d_1=r \cdot w, d_2=r \cdot h$), where $r$ is a random number between 0 and percentage $p \in 0.1 \cdot \{0, \dots, 5\}$, and $w$ and $h$ are the width and height of the image. When $p \in [0.1, 0.2]$, the model's performance is barely affected. However, a downward trend is observed when $p$ exceeds 0.3. This indicates that for shifts that do not impact the bbox class prediction, AOR is robust enough to produce correct predictions. Conversely, when the shift becomes larger, the model might misclassify the category during the first step of reasoning (e.g., a shifted left lung might be misclassified as the mediastinum), thereby affecting subsequent accuracy. This also indirectly demonstrates that our model performs step-by-step logical analysis of anatomical regions during reasoning.

\begin{table}[t]
\centering
\caption{Comparison of AOR's referring and grounding capabilities under different training strategy.}
\vspace{3mm}

\begin{tabular}{ccccccccc}
\toprule
\multicolumn{2}{c}{Strategy}& \multicolumn{3}{c}{VQA} & Referring & \multicolumn{3}{c}{Grounding} \\
Stage 1        & Stage 2  & verify & choose & query   & Acc      & R@0.3    & R@0.5   & R@0.7  \\ \hline
\ding{55}            & \ding{55}    &79.53 &67.96 &62.33       & 98.03      & 98.48     & 96.16     & 88.79         \\
\checkmark     & \ding{55}    &80.53 &70.05 &63.96    & 98.30      & 98.31      & 96.54     & 89.40    \\
\checkmark            & \checkmark   &\textbf{80.68}& \textbf{70.16} &\textbf{65.43}  & \textbf{98.58}    & \textbf{98.56}  & \textbf{96.62}  & \textbf{90.40}   \\ \bottomrule
\label{tab:diff stage}
\end{tabular}
\end{table}

\begin{figure}[t]
\centering
\includegraphics[width=\linewidth]{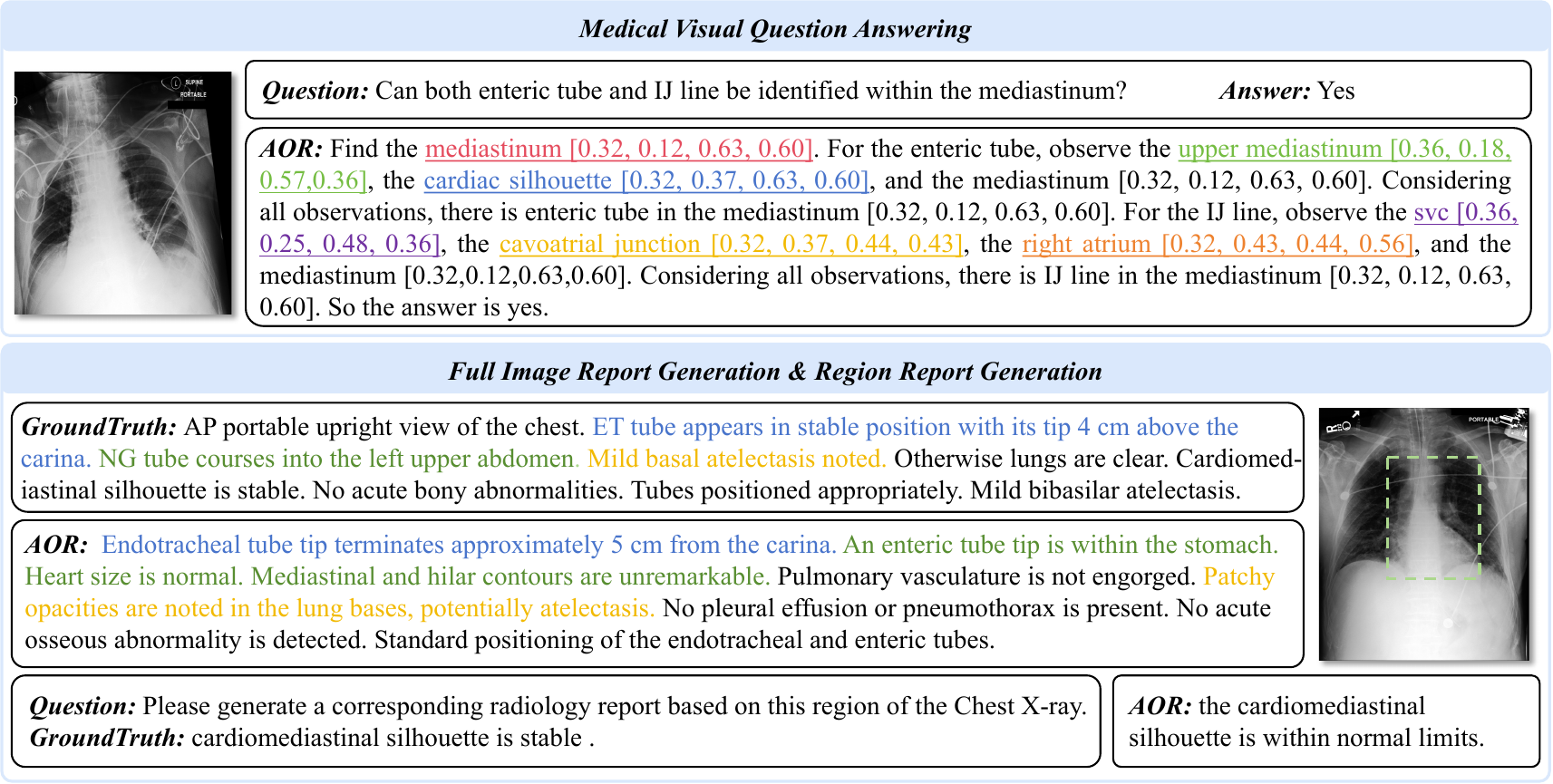}
\caption{Qualitative analysis of AOR on VQA task and report generation task.}
\label{fig:vis}
\end{figure}

\subsection{Qualitative Analysis}
Fig.~\ref{fig:vis} illustrates AOR's capabilities in medical grounded chat and referential dialogue. For the VQA task, AOR is capable of generating correct and logically reasoned answers. For the report generation task, due to the incorporation of fine-grained anatomical regions, AOR demonstrates a stronger grasp of details, such as ET tube, NG tube, and basal atelectasis. Moreover, it can generate corresponding report sentences for specified regions.

\section{Conclusion}
In this paper, we empower MLMMs with anatomy-centric reasoning capabilities, by (1) proposing the AOR framework, centers on the anatomical regions relevant to the given question, integrating the regions’ positional and representational information to conduct multimodal multi-step reasoning; (2) developing the medical CoT dataset AOR-Instruction, which provides tailored CoT answers for each VQA sample and strictly aligned region-sentence pairs for report generation. Experiments demonstrate the superiority of AOR over prior MLMMs in visual question answering, report generation, referring, and grounding, revealing its potential in clinical practice.

\bibliography{main}
\bibliographystyle{plain}

\end{document}